\documentclass[letterpaper, 10 pt, conference]{ieeeconf}

\IEEEoverridecommandlockouts

\usepackage{amsmath}
\usepackage{amssymb}
\usepackage[geometry]{ifsym}
\usepackage{cite}
\usepackage{graphicx}
\usepackage{tabularx}

\usepackage{ifsym}

\usepackage{caption}
\usepackage{subcaption}
\usepackage{array}
\usepackage[hidelinks]{hyperref}
\usepackage{algorithmic}
\usepackage[ruled,linesnumbered,noend]{algorithm2e}

\SetArgSty{textnormal}

\usepackage[normalem]{ulem}
\usepackage{color}
\usepackage[dvipsnames]{xcolor}

\newcommand{\new}[1]{{\textcolor{blue}{#1 }}}
\newcommand{\newc}[1]{{\textcolor{Green}{#1 }}}

\newcommand{\APbinding}{AP_{\psi}}
\newcommand{\APltl}{AP_{\varphi}}
\newcommand{\ltlpsi}{LTL$^\psi$}
\newcommand{\buchi}{B{\"u}chi }

\newcommand{\BuchiBeta}{\mathcal{\beta}^{t}_{j}}
\newcommand{\BuchiBetam}{\beta^{t}_{m}}

\newcommand{\syntline}{\: | \:}

\newcommand{\G}{\Box}
\newcommand{\F}{\Diamond}

\newcommand{\sigExT}{\sigma^{exT}}
\newcommand{\sigExF}{\sigma^{exF}}

\newcommand{\RN}[1]{%
  \textit{\uppercase\expandafter{\romannumeral#1}}%
}

\usepackage{amsthm}
\theoremstyle{plain}
\newtheorem{definition}{Definition}

\makeatletter
\let\NAT@parse\undefined
\makeatother
\usepackage{hyperref}  

\title{\bf
Online Resynthesis of High-Level Collaborative Tasks for Robots with Changing Capabilities
}
\author{Amy Fang, Tenny Yin, and Hadas Kress-Gazit
\thanks{Sibley School of Mechanical and Aerospace Engineering, Cornell University, Ithaca, NY, 14853 USA. {\tt\small \{axf4,yy389,hadaskg\}@cornell.edu}. This work is supported by the National Defense Science \& Engineering Graduate Fellowship 
(NDSEG) Fellowship Program.
}}

\begin{document}
\maketitle
\thispagestyle{empty}
\pagestyle{empty}

\begin{abstract} Given a collaborative high-level task and a team of heterogeneous robots and behaviors to satisfy it, this work focuses on the challenge of automatically, at runtime, adjusting the individual robot behaviors such that the task is still satisfied, when robots encounter changes to their abilities--either failures or additional actions they can perform.  We consider tasks encoded in \ltlpsi and minimize global teaming reassignments (and as a result, local resynthesis) when robots' capabilities change. We also increase the expressivity of \ltlpsi\ by including additional types of constraints on the overall teaming assignment that the user can specify, such as the minimum number of robots required for each assignment. We demonstrate the framework in a simulated warehouse scenario.
\end{abstract}

\section{Introduction}




There is a wealth of literature in planning for multi-robot systems due to its wide variety of applications, such as search and and rescue and warehouse automation. Recently, there has been a growing interest in using formal logic, such as Linear Temporal Logic (LTL), to capture multi-robot tasks with temporally extended tasks, such as surveillance and coverage, in a mathematically precise way (e.g.\cite{Ulusoy2012, Hustiu2020,Kloetzer2010, Kantaros2020, Chen2021}).

During execution, robots may encounter changes to themselves and to their environment; they may experience failures (e.g. a broken gripper) that limit what they can do, or they may acquire additional capabilities, such as a change in the environment (e.g. a new opening) that may allow them to reach previously unreachable areas. 
When considering multi-robot collaborative behavior, a single robot modification may affect the ability of the overall team to accomplish the task; as a result, other robots' behavior may need to change 
at runtime in order to successfully accomplish the task.

To account for such changes during execution, we extend our framework from \cite{Fang2024, Fang2024_TRO} that both automatically assigns robots to the task, as well as synthesizes high-level robot behaviors to satisfy a global task encoded in \ltlpsi. Here we propose a method for the team to autonomously adapt  when robot capabilities change  in the middle of execution while guaranteeing that the team is still able to satisfy the task. We aim to minimize the change in the original team assignment and behavior and only locally resynthesize a robot's behavior when possible; our approach only reconstructs the entire team when necessary. 

In addition to online resynthesis, we increase the expressivity of \ltlpsi \ to allow users to provide information on 1) the minimum number of robots that must be assigned to a specific subtask (captured through the notion of a \textit{binding}), and 2) which subtasks cannot be assigned to the same robot.

\textbf{Related Work:} 
Existing work have proposed methods to synthesize behavior for homogeneous multi-robot teams to satisfy temporal logic specifications \cite{Kloetzer2010,Kantaros2020,Yang2020}. For heterogeneous robots, the common approaches are either to decompose the global task into independent sub-tasks \cite{Faruq2018, Schillinger2018}, or \textit{a priori} task assignment \cite{Tumova2016, Verginis2024}. Approaches for heterogeneous teams to satisfy a global task include \cite{Sahin2017, Leahy2022, Luo2022}. The task is not explicitly decomposed; rather, portions of the task are assigned to robots based on their type or onboard capabilities. In our prior work \cite{Fang2024}, we proposed an extension of LTL, called \ltlpsi, in which a user can encode information about the relationships between actions and robots (e.g. the same robot that picks up a package must also drop it off). The synthesis framework was then extended in \cite{Fang2024_TRO} for \ltlpsi\ tasks to account for actions that take varying time duration to execute. In the aforementioned work, the task allocation happens offline prior to execution; resynthesis during execution is not considered. 

In  prior work \cite{Fang2022}, we considered resynthesis in scenarios where robots are already executing existing LTL tasks when new tasks are introduced. The distributed framework allows robots to resynthesize during execution such that they can interleave both tasks rather than perform them sequentially. Another application in which resynthesis is critical is in partially known or uncertain workspaces. In \cite{Guo2013}, robots revise their motion plan in real-time. The robots synthesize a preliminary motion plan, then iteratively revise the plan as the robot receives more information about its environment. 
\cite{Kalluraya2023} considers online revisions to robot plans for tackling reach-avoid problems encoded in LTL, specifically when environment is dynamic and uncertain. There also exists work that addresses the issue of robustness. For example, \cite{Ulusoy2012} generates plans online that are robust to timing errors. The work in \cite{Lindemann2021} introduces risk predicates to synthesize behavior that reduces the amount of risk in violating spatial temporal logic specifications.

To address resynthesis specifically due to robot failures, the approach in \cite{Zhou2022} first decomposes the global specification into independent sub-tasks, characterizes the disturbances into four types of failures, and uses a behavior tree to autonomously react to those failures. These failures are specific to quadruped and wheeled robots; they do not generalize to any type of robot. In \cite{Huang2022}, the authors consider a homogeneous team of robots executing a navigation task encoded in co-safe LTL. The user specifies at most how many robots can fail, and whenever a robot fails, the centralized planner
updates the global plan. The authors in\cite{Kalluraya2023} consider failures in capabilities, where each capability is a binary variable (either the robot has the capability or it does not). Our work considers a more granular level of failure in which a failure happens within a capability. A robot may no longer be able to execute specific actions within the capability (e.g. picking up an object with a robot manipulator), but other actions can still be executed (e.g. pushing an object with a robot manipulator). This approach allows us to address a broader range of potential failures that a may occur to a robot. In addition, we can consider other types of modifications to a robot's capability, such as a gripper being added during runtime or a change in the environment that increases the robot's reachable workspace. 

\textbf{Contributions:} In the context of synthesizing team and robot control from a high-level specification given in \ltlpsi,  we 1) increase the specification expressivity by allowing the user to provide constraints regarding the binding assignments (i.e. the minimum number of robots assigned to each binding and which bindings are allowed to be assigned together), and 2) propose a resynthesis framework for online adaptation to changes in robot capabilities. 
We demonstrate our approach in a simulated warehouse scenario.

\section{Task Grammar: \ltlpsi} \label{sec:ltlpsi}

We use \ltlpsi\ \cite{Fang2024,Fang2024_TRO} as the grammar for writing high-level collaborative tasks. In this work, we extend the grammar to allow additional constraints on the team composition. The task grammar for \ltlpsi is defined over atomic propositions that abstract robot actions, as well as 
bindings that relate actions to specific robots; 
any action associated with a given binding must be satisfied by all the robot(s) assigned that binding. Note that a robot may be assigned to multiple bindings, and a binding may be assigned to multiple robots.

An \ltlpsi\ specification $\varphi^{\psi}$ is defined recursively as: 
\begin{flalign}
    \psi &:= \rho \syntline \psi_1 \vee \psi_2 \syntline \psi_1 \wedge \psi_2 \\
     \varphi &:=  \pi  \ |  \ \neg\varphi  \ | \ \varphi \vee \varphi \ | \ \varphi \ \mathcal{U} \  \varphi \\
     \varphi^{\psi} \!&:= \varphi^{\psi} \! \syntline
     \neg(\varphi^{\psi}) \syntline
    \! \varphi_1^{\psi_1} \!\! \wedge \! \varphi_2^{\psi_2} | \
    \! \varphi_1^{\psi_1} \!\! \vee \! \varphi_2^{\psi_2} \! \syntline 
     \varphi_1^{\psi_1} \mathcal{U} \varphi_2^{\psi_2} \! \syntline 
     \G \varphi^{\psi}\hspace{-3mm}
\end{flalign} 


\noindent where $\psi$, the \textit{binding formula}, is a Boolean formula (excluding negation) over the binding propositions $\rho\in\APbinding$, and $\varphi$ is defined over the action propositions $\pi\in AP_{\varphi}$. 

In this work, we extend the expressivity of \ltlpsi \ by also defining the semantics over two types of binding constraints the user can now specify: 1) $c_{distinct}$, where $c \in c_{distinct}$ are sets of two or more bindings that cannot be allocated to the same robot (e.g. $\{\RN{1},\RN{2}\}\in c_{distinct}$ enforces ``a robot cannot be assigned both bindings \RN{1} and \RN{2}"), and 2) the set $c_{min}$, which contains the tuples $(\rho, k)$; this enforces that at least $k$ robots must be assigned binding $\rho$.


\textbf{Semantics: } The semantics of an \ltlpsi \ formula $\varphi^{\psi}$ 
are defined over 1) a team trace $\sigma = \sigma_1\sigma_2\ldots\sigma_n$, where $\sigma_j$ is the trace of robot $j$ such that $\sigma_j(i)$ is the set of atomic propositions $AP_{\varphi}$ that are true for robot $j$ at time step $i$; and 2) the team binding assignments $\mathcal{R} = \{r_{1}, r_{2},\ldots, r_{n}\}$, where $r_j \in \mathcal{R}$ is the set of bindings in $\APbinding$ that are assigned to robot $j$. 
For example, $r_{green} = \{\RN{1}\}, r_{blue} = \{\RN{1},\RN{2}\}$ indicates that green robot is assigned binding \RN{1}, and the blue robot is assigned bindings \RN{1} and \RN{2}. We also define the function $\zeta: \psi \rightarrow 2^{2^{\APbinding}}$, which outputs all possible combinations of $\rho \in \APbinding$ that satisfy $\psi$. For example, $\zeta \bigl( \RN{1} \vee (\RN{2} \wedge \RN{3}) \bigr) = \{ \{\RN{1}\},\{\RN{2},\RN{3}\}, \{\RN{1},\RN{2},\RN{3}\}\}$. 

Given $c_{distinct}$ and $c_{min}$, and given the semantics of LTL~\cite{Baier2008} we define the semantics of \ltlpsi\ as follows:

\begin{itemize}
\setlength\itemsep{0.6em}
    \item $(\sigma(i), \mathcal{R})\! \models \varphi^{\psi} \text{ iff } \exists K \in \zeta(\psi)
    \text{ s.t. } (K \subseteq \bigcup\limits_{p=1}^{n} r_{p}) \\ \text{ and } (\forall j \text{ s.t. } K \cap r_j \neq \emptyset, \sigma_j(i) \models \varphi) \\ \text{ and }
    (\forall r_j \in \mathcal{R}, \forall c \in c_{distinct}, \ c \not\subseteq r_j)  \\
    \text{ and } (\forall (\rho, k) \in c_{min}, \ \lvert \{r_j \in \mathcal{R} \ | \ \rho \in r_j\} \rvert \geq k)$ 

    \item $(\sigma(i), \mathcal{R})\! \models (\neg \varphi)^{\psi} \text{ iff } \exists K \in \zeta(\psi)
    \text{ s.t. } (K \subseteq \bigcup\limits_{p=1}^{n} r_{p}) \\ \text{ and } (\forall j \text{ s.t. } K \cap r_j \neq \emptyset, \sigma_j(i) \not\models \varphi) \\ \text{ and }
    (\forall r_j \in \mathcal{R}, \forall c \in c_{distinct},  \ c \not\subseteq r_j)  \\
    \text{ and } (\forall (\rho, k) \in c_{min}, \ \lvert \{r_j \in \mathcal{R} \ | \ \rho \in r_j\} \rvert \geq k)$ 

    \item $(\sigma(i), \mathcal{R})\! \models \neg(\varphi^{\psi}) \text{ iff } \exists K \in \zeta(\psi)
    \text{ s.t. } (K \subseteq \bigcup\limits_{p=1}^{n} r_{p}) \\ \text{ and } (\exists j \text{ s.t. } K \cap r_j \neq \emptyset, \sigma_j(i) \not\models \varphi) \\ \text{ and }
    (\forall r_j \in \mathcal{R}, \forall c \in c_{distinct},  \ c \not\subseteq r_j)  \\
    \text{ and } (\forall (\rho, k) \in c_{min}, \ \lvert \{r_j \in \mathcal{R} \ | \ \rho \in r_j\} \rvert \geq k)$ 
    
    \item $(\sigma(i),\!\mathcal{R})\!\!\models\!\! \varphi_1^{\psi_1} \!\wedge\varphi_2^{\psi_2}$ iff $(\sigma(i),\!\mathcal{R})\!\!\models \! \varphi_1^{\psi_1}$and $(\sigma(i),\!\mathcal{R})\! \models\!\varphi_2^{\psi_2}$
    
    \item $(\sigma(i),\!\mathcal{R})\! \models\! \varphi_1^{\psi_1}\!\vee\! \varphi_2^{\psi_2}$ iff $(\sigma(i),\!\mathcal{R})\! \models\!\varphi_1^{\psi_1}$or $(\sigma(i), \mathcal{R})\! \models \varphi_2^{\psi_2}$
    

    
    

    \item $(\sigma(i), \mathcal{R}) \models \varphi_1^{\psi_1} \ \mathcal{U} \ \varphi_2^{\psi_2}$ iff $\exists \ell \geq i$ s.t. $(\sigma(\ell), \mathcal{R}) \models \varphi_2^{\psi_2}$ and $\forall i \leq k < \ell, (\sigma(k), \mathcal{R}) \models \varphi_1^{\psi_1}$
    
      \item $(\sigma(i), \mathcal{R}) \models \G \varphi^{\psi}$ iff $\forall \ell>i, (\sigma(\ell), \mathcal{R}) \models \varphi^{\psi}$


\end{itemize}



Intuitively, a team of robots satisfies the formula $\varphi^{\psi}$ if and only if the following conditions hold: 1) there exists a set of bindings $K \in \zeta(\psi)$ for which all the bindings are assigned to (at least one) robot; 2) for all robots assigned these bindings, their traces satisfy $\varphi$~\cite{Baier2008}; 3) none of the robots are assigned any combinations of bindings $c \in c_{distinct}$; 
any binding not appearing in $c_{distinct}$ must still be assigned to at least one robot; and 4) at least $k$ number of robots are assigned binding $\rho$ for every $(\rho, k) \in c_{min}$.

\textit{Example: }
\begin{align} \label{eq:ex1}
\begin{split}
  \varphi^{\psi} = &\ \F(beep \wedge storage_c)^{\RN{3}} \wedge \F dock_c^{\RN{1}} \\
&\wedge \G(dock_c^{\RN{1}} \rightarrow (roomB_c \wedge camera )^{\RN{2}})  \\
&c_{distinct} = \ \emptyset, \ c_{min} = \{(\RN{1}, 2)\}
\end{split}
\end{align}

In English, the task captures ``all robots assigned binding \RN{3} must go to the storage room and beep, and all robots assigned binding \RN{1} must eventually go to the dock. Anytime all the robots assigned binding \RN{1} are in the dock, all robots assigned binding \RN{2} must be in room B and taking a picture. At least two robots must be assigned binding \RN{1}." 


\section{Prior work - Robot Model and \buchi Automaton}

\subsection{Robot Model}

Each robot $j$ is modeled according to its set of capabilities, $\Lambda_j =  \{\lambda_1, \ldots, \lambda_k\}$\cite{Fang2022}. Each capability is a transition system $\mathcal{\lambda} = (X, x_0, AP, \Delta, \mathcal{L}, \mathcal{W})$, where $X$ is a set of states, $x_0 \in X$ is the initial state, $AP$ is the set of atomic propositions that are an abstraction of the actions the capability can execute, $\Delta \subseteq X \times X$ is a transition relation,  $\mathcal{L}: X \rightarrow 2^{AP}$ is the labeling function, and $\mathcal{W}: \Delta \rightarrow \mathbb{R}_{\ge 0}$ is the cost function; each transition is assigned a weight using $\mathcal{W}$.

A robot model $A_j$\cite{Fang2022} is the product of its capabilities: $A_j = \lambda_1 \times\ldots \times \lambda_k$ such that $A_j=(S, s_0, AP_j, \gamma, L, W)$. $S = X_1 \times \ldots \times X_k$ is the finite set of states, $s_0 \in S$ is the initial state, $AP_j = \bigcup_{i=1}^k AP_i$
is the set of propositions, $\gamma \subseteq S \times S$ is the transition relation, $L:S \rightarrow 2^{AP_j}$ is the labeling function, and $W: \gamma \rightarrow \mathbb{R}_{\ge 0}$ is the cost function. 
The constraints of the workspace the robot operates in are encoded in
its motion capability.

\subsection{\buchi Automaton for an \ltlpsi\ Formula}\label{sec:buchi_ltlpsi}

An LTL formula $\varphi$ can be translated into a Nondeterministic \buchi automaton $\mathcal{B}= (Z, z_0, \Sigma_{\mathcal{B}}, \delta_{\mathcal{B}}, F)$, where $Z$ is the set of states, $z_0 \in Z$ is the initial state, $\Sigma_{\mathcal{B}}$ is the input alphabet, $\delta_{\mathcal{B}}: Z \times \Sigma_{\mathcal{B}} \times Z$ is the transition relation, and $F \subseteq Z$ is the set of accepting states.
An infinite run of $\mathcal{B}$ over a word $\sigma = \sigma_1 \sigma_2 \sigma_3 \ldots \in \Sigma_{\mathcal{B}}$ is an infinite sequence of states $\mathcal{Z} = z_0 z_1 z_2\ldots$ such that $(z_{i-1}, \sigma_i, z_{i}) \in \delta_{\mathcal{B}}$. A run is accepting if and only if an accepting state or set of accepting states appear infinitely often in $\mathcal{Z}$, i.e. Inf($\mathcal{Z}$) $\cap \ F \neq \emptyset$ \cite{Baier2008}.

When creating a \buchi automaton for an \ltlpsi\ formula, we first rewrite the formula to include only propositions of the form $\pi^\rho$~\cite{Fang2024}; then {$\Sigma_{\mathcal{B}} = 2^{AP_{\varphi}^{\psi}}\times 2^{AP_{\varphi}^{\psi}}\times 2^{AP_{\varphi}^{\psi}}\times 2^{AP_{\varphi}^{\psi}}$} and $\sigma=(\sigma^T,\sigExT,\sigma^F, \sigExF)\in\Sigma_{\mathcal{B}}$. $\sigma^T$ and $\sigma^F$ are the sets of propositions $\pi^\rho$ that are true/false for all robots; $\sigExT$ and $\sigExF$ are the sets of propositions $\pi^\rho$ that are true/false for at least one robot. 
The set of $\sigma^T \cup \sigma^F$ are denoted as \textit{for all} propositions, and $\sigExT \cup \sigExF$ as \textit{there exists} propositions. 


\section{Behavior Synthesis}
To synthesize robot behavior, we take the product of the robot model and the \buchi automaton and find a satisfying trace~\cite{Fang2024_TRO}.


\begin{definition}[Capability  Function]
$\mathfrak{C}: \Sigma_{\mathcal{B}} \times \APbinding \rightarrow 2^{\APltl} \times 2^{\APltl} \times 2^{\APltl} \times 2^{\APltl}$ such that for $(\sigma^T,\sigExT,\sigma^F, \sigExF) \in \Sigma_{\mathcal{B}}, \rho \in \APbinding$, $  \mathfrak{C}(\sigma, \rho) = (C_T, C_{exT}, C_F, C_{exF})$, where for $k \in \{T, exT, F, exF\}$, $C_k = \{\pi \in\APltl \ | \ \exists \pi^\rho \in \sigma^k\}$.
\end{definition}

Given a binding $\rho$, $C_T$ and $C_F$ are the sets of propositions $\pi$ in which $\pi^\rho$ is a \textit{for all} proposition that is True/False and appear with binding $\rho$ in label $\sigma$ of a \buchi transition; $C_{exT}$ and $C_{exF}$ are defined similarly for \textit{there exists} propositions with binding $\rho$. For example, the transition between states 3 and 0 of the \buchi automaton in Fig. \ref{fig:buchi} is $\sigma = \{ \{roomB_c^\RN{2}, camera^\RN{2}, dock_c^\RN{1}\}, \emptyset, \emptyset, \emptyset \}$. Then, 
$\mathfrak{C}(\sigma, \RN{2}) = (\{roomB_c, camera\},\emptyset, \emptyset, \emptyset )$. 

We modify the following definition from \cite{Fang2024_TRO} to account for the user-specified constraint $c_{distinct}$: 
\begin{definition}[Binding Assignment Function]
Given two states in the robot model, $s$ and $s'$, and
$\sigma = (\sigma^T,\sigExT,\sigma^F, \sigExF)$, 
$\mathfrak{R}(s, \sigma, s') = \{r \in 2^{\APbinding} \setminus \emptyset \ | \ \forall c \in c_{distinct}, c \nsubseteq r$ and $\forall \rho \in r, \mathfrak{C}(\sigma, \rho)=(C_T, C_{exT}, C_F, C_{exF}),  \bigcup_{\rho \in r} (C_T \cup C_{exT}) \subseteq L(s')$ and $\bigcup_{\rho \in r} (C_F \cup C_{exF}) \cap L(s') = \emptyset \}$. 
\end{definition}

The output of function $\mathfrak{R}$ is the set of all combinations of binding propositions that can be assigned to a robot over a given transition $\sigma$ in the \buchi automaton. A robot can be assigned a set of binding propositions $r$ if and only if the following are satisfied: {1) $r$ is not a superset of any set in $c_{distinct}$, which are  combinations of bindings that cannot be assigned to the same robot}; 2) for all $\rho \in r$, all propositions $\pi$ that are in $\sigma^T \cup \sigma^{exT}$ as $\pi^\rho$ also appear in the state label of the next state $s'$; and 3) for all $\rho \in r$, none of the propositions $\pi$ that appear in $\sigma^F \cup \sigma^{exF}$ as $\pi^\rho$ also appear in the state label of $s'$.



To synthesize behavior for a robot~\cite{Fang2024_TRO}, we find the minimum cost accepting trace in its product automaton 
$\mathcal{G}_j = A_j \times \mathcal{B} = (Q, q_0, AP_j,\delta_\mathcal{G},L_\mathcal{G}, W_\mathcal{G}, F_\mathcal{G})$, where 

\begin{itemize}
    \item $Q = S \times Z$ is a finite set of states
    \item $q_0 = (s_0, z_0) \in Q$ is the initial state
    \item $\delta_\mathcal{G}\subseteq Q\times Q$ is the transition relation, where for $q = (s, z)$ and $q' = (s',z')$, $(q,q')\in \delta_\mathcal{G}$ if and only if $(s,s') \in \gamma$ and $\exists \sigma \in \Sigma_{\mathcal{B}}$ such that $(z, \sigma, z') \in \delta_{\mathcal{B}}$ and $\mathfrak{R}(q, \sigma, q') \neq \emptyset$ 
    


    \item $L_\mathcal{G}$ is the labeling function s.t. for $q = (s, z)$, $L_\mathcal{G}(q)\!=\!L(s)\!\subseteq\!AP_j$ 
    \item $W_\mathcal{G}: \delta_{\mathcal{G}} \rightarrow \mathbb{R}_{\ge 0} $ is the cost function s.t. for $(q, q') \in \delta_{\mathcal{G}}$, $q = (s, z)$, $q' = (s',z')$, $W_\mathcal{G}((q, q'))=W((s, s'))$
    \item $F_\mathcal{G} = S \times F$ is the set of accepting states
\end{itemize}

If a team of robots and their synthesized behavior follow the same trace in the \buchi automaton $\mathcal{B}$ to an accepting cycle, the team is guaranteed to satisfy the task. We denote such a collective trace as $\beta$. 

\section{Problem Setup}

\subsection{Modifications}

\begin{figure}[t!]
     \centering
     \begin{subfigure}{\columnwidth}
         \centering
         \includegraphics[width=0.85\textwidth]{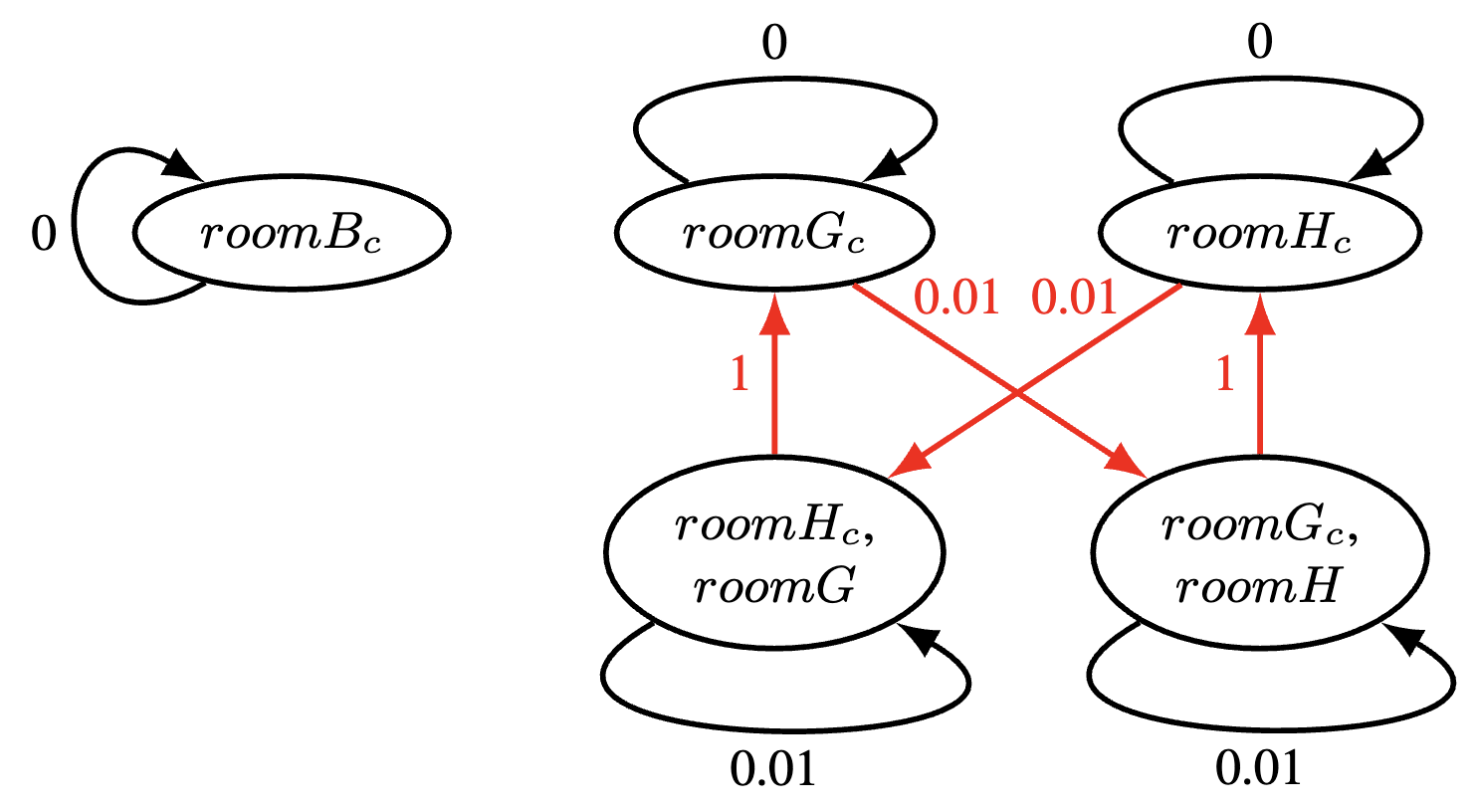}
         \caption{}
         \label{fig:cap_fail}
     \end{subfigure}
    \centering
     \begin{subfigure}{\columnwidth} 
         \centering
         \includegraphics[width=0.85\textwidth]{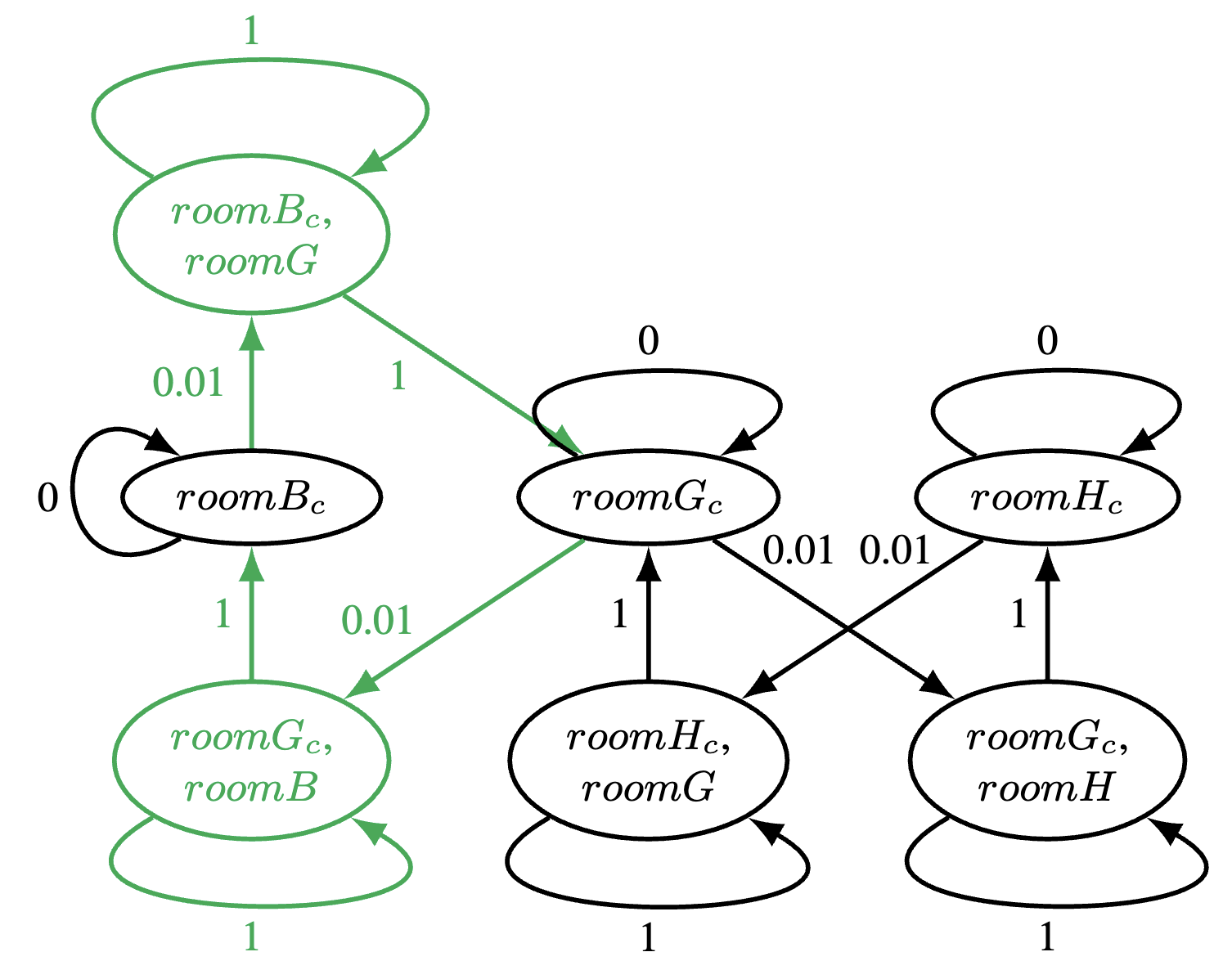}
         \caption{}
         \label{fig:cap_aug}
     \end{subfigure}
    \caption{Modifications to $\lambda_{mot}$ in which (a) the robot can no longer move between rooms G and H ($\Delta_{mot}^{rem} $ is the set of red transitions), and (b) the robot can now move between rooms B and G ($\Delta_{mot}^{add} $ is the set of green transitions). 
    }
    \label{fig:cap_mod}
\end{figure}

We define a robot capability modification as a change in the capability's transition relation $\Delta$. {Specifically, this involves either adding new transitions to $\Delta$ or removing existing ones. For robot $m$, we define $\Delta_m^{add}$ and $\Delta_m^{rem}$ where} 
{$\Delta_m^{add} = \{\Delta_{m,a}^{add},\Delta_{m,b}^{add}, \ldots\}$, $\Delta_{m,\alpha}^{add}$ is the set of transitions added to capability $\lambda_\alpha$; $\Delta_m^{rem}= \{\Delta_{m,a}^{rem},\Delta_{m,b}^{rem}, \ldots\}$, $\Delta_{m,\alpha}^{rem}$ is the set of transitions that are removed from capability $\lambda_\alpha$.}
{The cost function of a capability $\mathcal{W}$ may also change, particularly if transitions are added. We represent the corresponding set of cost functions as $\mathcal{W}_m^{add}$ and $\mathcal{W}_m^{rem}$.}
Fig. \ref{fig:cap_mod} shows examples of  modifications to a robot's motion capability $\lambda_{mot}$.



\subsection{Problem Statement}

Given a team of heterogeneous robots ${A}$ 
executing $\varphi^{\psi}$ with binding assignments $\mathcal{R}_{{A}}$, and given the sets of capability modifications to robot $m$, $\Delta_m^{add}, \Delta_m^{rem}$, and the corresponding set of cost functions $\mathcal{W}_m^{add}, \mathcal{W}_m^{rem}$, find a (possibly) new assignment $\mathcal{R}_{A}^{mod}$ and trace $\sigma^{mod}$ such that $(\sigma^{mod},\mathcal{R}_{A}^{mod})\models \varphi^{\psi} $. 

We assume that at any instance, only one robot is modified. We also assume each robot is aware of its modifications when they happen, and that robots have all-to-all communication with one another; this is to facilitate the binding reallocation and synchronization processes. 

\subsection{Example} \label{sec:ex}

\begin{figure}[h!]
    \centering 
    \includegraphics[width=0.65\columnwidth]{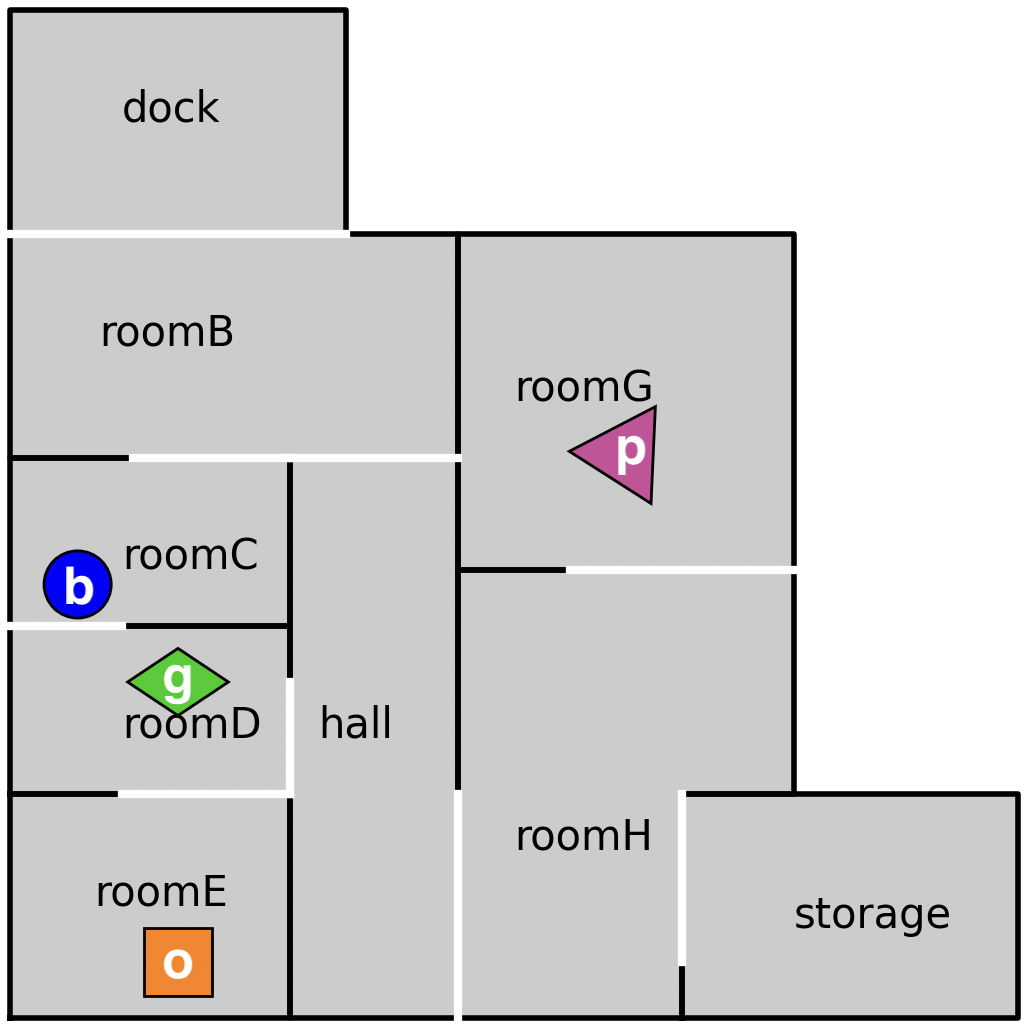}
    \caption{Environment and robot setup}
    \label{fig:setup}
\end{figure}

Consider a team of robots $A = \{A_{green}, A_{blue}, A_{orange},$ $ A_{pink}\}$ in a warehouse environment shown in Fig. \ref{fig:setup}. The robots’ capabilities and labels on their initial state are:
\begin{align*}
    \Lambda_{green} &= \{\lambda_{\mathit{mot}}, \lambda_{camera}\} &L(s_0)& = \{roomD_c\} \\
    \Lambda_{blue} &= \{\lambda_{\mathit{mot}}, \lambda_{beep}\} &L(s_0)& = \{roomC_c\} \\
    \Lambda_{orange} &= \{\lambda_{\mathit{mot}}\} &L(s_0)& = \{roomE_c\} \\
    \Lambda_{pink} &= \{\lambda_{\mathit{mot}},  \lambda_{beep},\lambda_{cam}, \lambda_{scan}\} &L(s_0)& = \{roomG_c\}
\end{align*}

\begin{figure*}[h!]
    \centering 
    \includegraphics[width=0.9\textwidth]{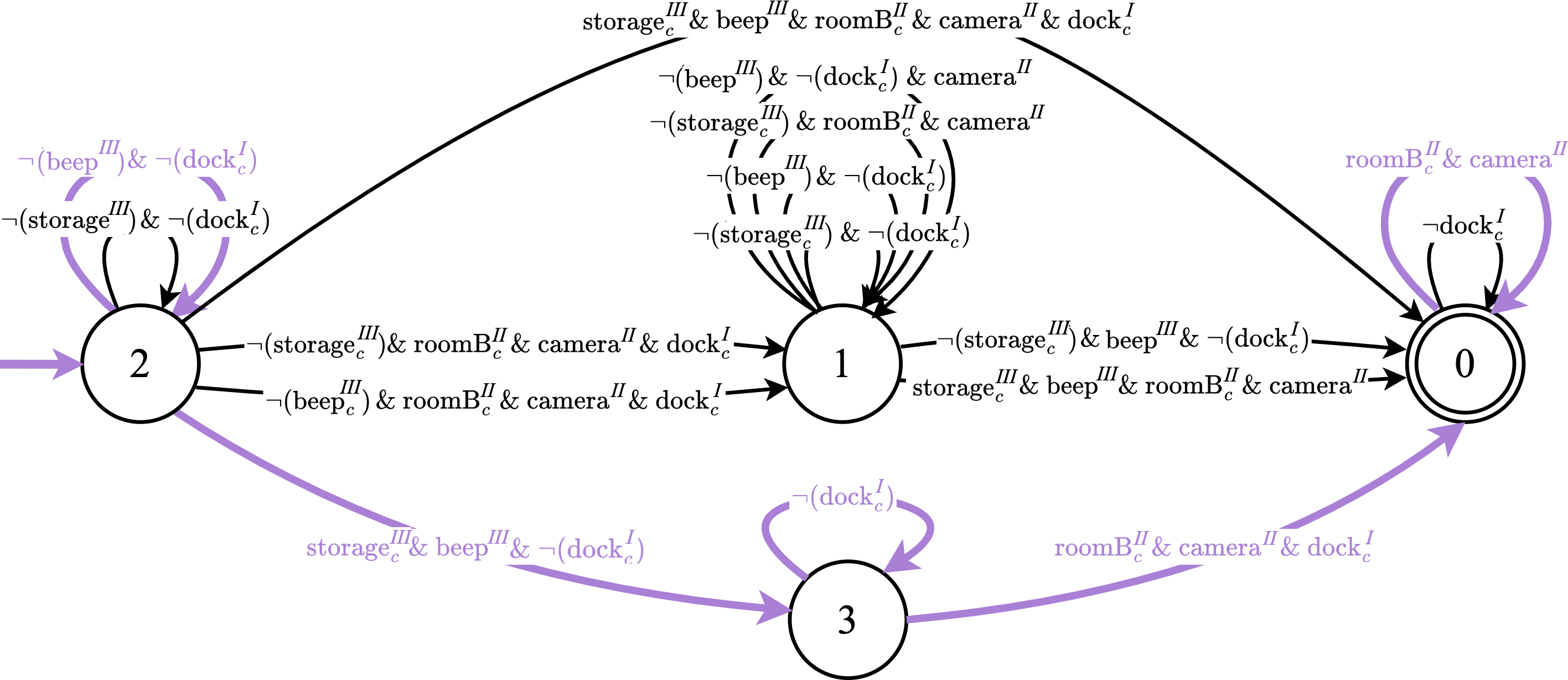}
    \caption{Buchi Automaton $\mathcal{B}$ for the example in Sec. \ref{sec:ex}. The highlighted transitions is $\beta$, the trace that the team of robots are collectively traversing to satisfy the task.}
    \label{fig:buchi}
\end{figure*}

The robots are currently executing the task in Eq. \ref{eq:ex1} with binding assignments $\mathcal{R}_{{A}} = \{r_{green}, r_{blue}, r_{orange}, r_{pink}\}$, where $r_{green} = \{\RN{1}\}$, $r_{blue} = \{\RN{1}, \RN{3}\}$, $r_{orange} = \{\RN{1}\}$, $r_{pink} = \{\RN{2},\RN{3}\}$. The trace $\beta$ in the \buchi automaton that the robots collectively satisfy is shown in purple in Fig. \ref{fig:buchi}. $\mathcal{R}_{{A}}$ and $\beta$ are automatically found using the method proposed in \cite{Fang2024_TRO}.


During execution, modifications to the robots' capabilities occur. In the following sections, we illustrate how the robots conduct online resynthesis such that the overall team still satisfies the task.

\section{Approach: Resynthesis Framework}

\begin{figure}[h!]
    \centering 
    \includegraphics[width=\columnwidth]{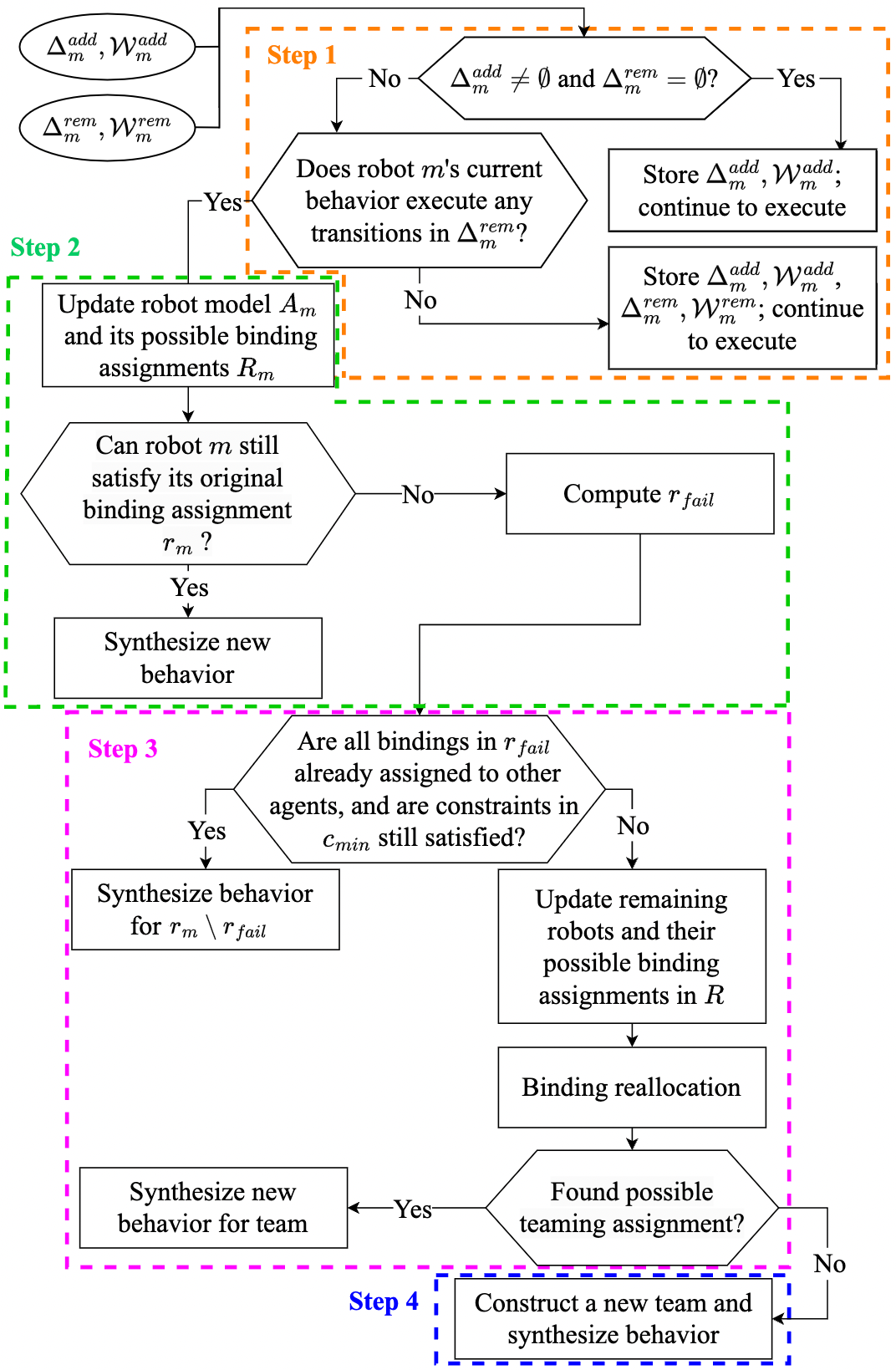}
    \caption{Overview of resynthesis process}
    \label{fig:flowchart}
\end{figure}

We update the behavior of the robots on the fly when capability modifications occur. A summary of the overall framework is shown in Fig. \ref{fig:flowchart}. We only synthesize new behavior when necessary; Section \ref{sec:flow} describes how we determine whether resynthesis is needed. 
Section \ref{sec:upd_prod_aut} outlines the method for a robot to update its product automaton. Section \ref{sec:allocation} outlines the binding reassignment process.

\subsection{Evaluating Modified Robot's Behavior} \label{sec:flow}

When robot $m$ is modified (i.e. $\Delta_m^{add} \cup \Delta_m^{rem} \neq \emptyset$), it first determines whether its modifications will affect the ability of the team to successfully execute the task. It does so by executing the following steps:

\textbf{Step 1}:
First, the robot checks if $\Delta_m^{add} \neq \emptyset$ (the robot has expanded its capabilities)  and $\Delta_m^{rem} = \emptyset$ (the robot has not lost any of its capabilities).
In this case, 
the robot is still able to execute its original behavior. Thus, this information is stored and then incorporated when binding reassignment is necessary; the robot continues executing its original behavior. When $\Delta_m^{rem} \neq \emptyset$, i.e. transitions are removed from a robot's capability, the robot checks if its original behavior contains any of those transitions. 
If not, the robot continues to execute its original behavior 
and stores these modifications. 

\textbf{Step 2}: The robot proceeds to step 2 if its original behavior includes a transition it can no longer perform, i.e. {for behavior $b_m = q_1q_2 \ldots$, where $q_i=(s_i, z_i)$, $\exists i \in \{2, \ldots, |b_m|\} \text{ and } \exists (x,x') \in \bigcup_{a} \Delta_{j,a}^{rem} \text{ s.t. } x \in s_{i-1}$, $x' \in s_i\}$}. 
It updates its model and product automaton to reflect the modifications (Section~\ref{sec:upd_prod_aut}).

Based on its updated robot model and product automaton, it determines if there are any bindings it was originally assigned, $r_m$, that it can no longer do. These bindings are stored in $r_{fail}$.

The team of robots may have overlapping binding assignments. Thus, if all bindings in $r_{fail}$ are already assigned to other robots and the user-specified constraints regarding the minimum number of robots assigned to each binding $c_{min}$ are still satisfied, only robot $m$ needs to resynthesize a behavior for the remaining bindings it can do ($r_m \setminus r_{fail}$); the rest of the team continues executing their original behavior.

\textbf{Step 3}: If there are bindings in $r_{fail}$ that are not already assigned to other robots, then binding reallocation is necessary. Before this can happen, each robot $j$ updates $R_j$, the set of all possible binding combinations it can do. To do so, each robot updates its product automaton based on any modifications that have been previously stored (see Sec. \ref{sec:upd_prod_aut}). 
Updating of each robot's model and product automaton is done in a distributed manner; only the binding reallocation algorithm is centralized. {During reallocation, we minimize the number of robots that change assignments, since every reassignment requires the robot to resynthesize its behavior}.

\textbf{Step 4}: In the worst case scenario, if there is no possible binding reallocation such that the robots can satisfy the task, we synthesize a new team using the framework proposed in \cite{Fang2024_TRO} to find another collective trace in the \buchi automaton.

\subsection{Updating the Product Automaton} \label{sec:upd_prod_aut}

Before resynthesizing their behavior, each robot $j$ updates its model and product automaton based on current and past modifications, if any. 
We update the product automaton $\mathcal{G}_j$ based on the sets of transitions to be added $\Delta_j^{add}$ and removed $\Delta_j^{rem}$. To do so, we construct $\BuchiBeta$, which represents the portion of the collective trace $\beta$ that has not yet been executed by robot $j$ at the step $t$ when the modification is introduced, and $A_j^{mod}$, which contains only the affected transitions of the robot model. This way, we can remove or add the  transitions to the product automaton based on $\mathcal{G}_j^{mod} = \BuchiBeta \times A_j^{mod}$, rather than reconstructing the entire product automaton of a robot $\mathcal{G}_j=\mathcal{B} \times A_j$ from scratch. 



\textbf{Constructing $\BuchiBeta$ : } 
Since we know the trace $\beta$ in the \buchi automaton that the team is traversing, we only need to check how the capability modification affects the transitions in $\beta$.

Let the modification to robot $m$ occur when it is at $q_m^t = (s_m^t, z_m^t)$ in its behavior. Then, for each robot $j$, we define $\BuchiBeta$ as the reachable portion of $\beta$ from $z_j^t$. Note that $\BuchiBeta$ may not be equivalent $\BuchiBetam$; that is, given a robot $j$ and the state it is at when the modification occurs, $q_j^t = (s_j^t, z_j^t)$, $z_j^t$ may not be equal to $z_m^t$ for any robot $j \neq m$ due to the synchronization policy each robot executes, as outlined in \cite{Fang2024}: For each transition $(z, \sigma, z') \in \beta$, where $z \neq z'$, robots that are assigned bindings that appear in $\sigma$ must wait to execute their transition $((s, z), (s', z'))$ until all other relevant robots are also ready. Thus, any robots whose assigned bindings do not appear in $\sigma$ are not involved in the synchronization requirement. For example, given the \buchi automaton shown in Fig. \ref{fig:buchi}, if a robot's assigned bindings are $r=\{\RN{2}\}$, then it does not need to wait to synchronize for transition $(2, \sigma, 3) \in \beta$, where $\sigma = \{\{storage_c^\RN{3}, beep^\RN{3}\}, \emptyset, \emptyset, \{dock_c^\RN{1}\}\}$, since binding \RN{2} does not appear on any proposition in $\sigma$. 

Although the robots may be at different states in $\beta$ when the modification occurs, we want to ensure that the entirety of $\beta$ is still satisfied when the robots resynthesize new behavior. To do so, let $\textsc{Index}(\beta, q_j^t) = \{i \in \{1, \ldots,|\beta|\}\:|\: \beta[i] = z_j^t\}$, where $q_j^t = (s_j^t, z_j^t)$. This outputs the index of state $z_j^t$ in trace $\beta$.

If $\textsc{Index}(\beta, q_j^t) > \textsc{Index}(\beta, q_m^t)$, then robot $j$ is ``ahead" of robot $m$ in the trace $\beta$, implying that it did not have to participate in any synchronization policies between states $z_m^t$ and $z_j^t$ in $\beta$. This means that robot $j$ can be in any state in its robot model without violating any transitions (and states) between $z_m^t$ and $z_j^t$. Thus, to guarantee that the entire trace $\beta$ is satisfied, we take the conservative approach and move robot $j$ ``back" to $z_m^t$ 
; i.e. we modify the state from $q_j^t = (s_j^t, z_j^t)$ to $q_j^t = (s_j^t, z_m^t)$ and $\BuchiBeta = \BuchiBetam$. Conversely, $\textsc{Index}(\beta, q_m^t) \geq \textsc{Index}(\beta, q_j^t)$ indicates that the modified robot $m$ is ``ahead" of robot $j$ in $\beta$ and can maintain any truth value between $z_j^t$ and $z_m^t$. Thus, $q_j^t$ and $\BuchiBeta$ remain unchanged.

\textbf{Constructing $A_j^{mod}$: } 
The approach to constructing $A_j^{mod}$, the  robot model with the modified transitions, differs depending on if the transitions in the robot capabilities need to be removed ($\Delta_j^{rem}$) or added ($\Delta_j^{add}$).

\subsubsection{Constructing $A_j^{mod}$ with $\Delta_j^{rem}$}
$\Delta_{j,a}^{rem} \in \Delta_j^{rem}$ is the set of transitions $(x, x')$ that are no longer valid in the robot's capability $\lambda_a$, and we need to remove the transitions $((s,z), (s',z')) \in \delta_\mathcal{G}$ in the robot's product automaton $\mathcal{G}_j$ that are no longer valid. To do so, we first construct the robot model with the modified transitions, $A_j^{mod} = (S_j^{rem}, s_j^t, AP_j, \gamma_j^{rem}, L, W)$, where $S_j^{rem} \subseteq S, \gamma_j^{rem} \subseteq \gamma$ are defined as
\begin{align}
    \!\!\!\!\gamma_j^{rem} \!=\! \{(s, s') \!\!\in \!\gamma_j \ \!\! &| \ \!\exists (x, x') \!\in\! \bigcup_{a} \Delta_{j,a}^{rem} \!\text{ s.t. }\! x \!\in\! s, x' \!\in\! s' \}\\
    &S_j^{rem} = \bigcup_{(s,s') \in \gamma_j^{rem}} s \cup s'
\end{align}







\subsubsection{Constructing $A_j^{mod}$ with $\Delta_j^{add}$}
$\Delta_{j,a}^{add} \in \Delta_j^{add}$ is the set of transitions $(x, x')$ to be added to the robot's existing capability $\lambda_a$, 
and $\mathcal{W}_{j,a}^{add}$ is the set of cost functions that assigns a weight to the added transitions. Note that $x$ or $x'$ might be new states in the capability.

Let the current robot model be $A_j = \lambda_a \times \cdots \times \lambda_k$. Without loss of generality, let $\Delta_{j,a}^{add}$, $\mathcal{W}_{j,a}^{add}$ be the set of transitions and cost function, respectively, in capability $\lambda_a$ that is being added, where $\lambda_a = (X_a, x_a^t, AP_a, \Delta_a, \mathcal{L}_a, \mathcal{W}_a)$. Then, $\lambda_a^{add} = (X_a^{add}, x_a^t, AP_a, \Delta_{j,a}^{add}, \mathcal{L}_a, \mathcal{W}_{j,a}^{add})$, where $X_a^{add} = \bigcup_{(x,x')\in \Delta_{j,a}^{add}} \{x, x'\}$. The added portion of the robot model is $A_j^{mod} = \lambda_a^{add} \times \cdots \times \lambda_k$.

\textbf{Constructing $\mathcal{G}_j^{mod}$: }
Using $\BuchiBeta$ and $A_j^{mod}$, we construct the affected product automaton $\mathcal{G}^{mod}_j = A^{mod}_j \times \BuchiBeta$.

If we are considering $\Delta_j^{rem}$, then the portion of the product automaton that is to be removed is $\mathcal{G}_j^{mod} = \BuchiBeta \times A_j^{mod}$. We modify the original product automaton by removing the transitions $\delta_\mathcal{G}^{mod}$, i.e. $\delta_\mathcal{G} = \delta_\mathcal{G} \setminus \delta_\mathcal{G}^{mod}$.

If we are considering $\Delta_j^{add}$,  then we add $\mathcal{G}_j^{mod}$ to the original product automaton, i.e. $\delta_\mathcal{G} = \delta_\mathcal{G} \cup \delta_\mathcal{G}^{mod}$, $Q = Q \cup Q^{mod}$, and 
\begin{equation*}
W_\mathcal{G}((x,x')) = 
    \begin{cases}
        W_\mathcal{G}((x,x')) &  (x,x')\in \delta_\mathcal{G} \setminus \delta_\mathcal{G}^{mod}\\
        W_\mathcal{G}^{mod}((x,x')) &  (x,x')\in  \delta_\mathcal{G}^{mod}\\
    \end{cases}  
\end{equation*}

\noindent where $\delta_\mathcal{G}^{mod}$, $Q^{mod}$, $W_\mathcal{G}^{mod}$ are the transitions, states, and cost function, respectively, in $\mathcal{G}^{mod}_j$.





\subsection{Binding (Re)Allocation} \label{sec:allocation}
We modify the binding allocation framework proposed in~\cite{Fang2024} such that it 1) allows users to provide constraints on the minimum number of robots that must be assigned to a specific binding or which bindings are not allowed to be assigned to the same robot (Section \ref{sec:ltlpsi}, shown in green in Alg. \ref{algo:allocation}), and 2) to reallocate robots to bindings in response to modifications such that we minimize the number of robots that are reassigned different bindings (shown in blue in Alg. \ref{algo:allocation}). 

Given the set of robots and $R$, where $R_j \in R$ is set of all possible binding assignments robot $j$ can do, 
the goal is to assign each binding to the minimum number of robots it requires (Alg. \ref{algo:allocation}).
We intialize the set of unassigned bindings, $unassigned$, to be the set of all bindings $\mathbf{b}$. For each round of binding allocation, the framework selects the robot $j$ to be assigned based on the following ordering:

\begin{enumerate}
    \item Given the set of unassigned robots, we first choose the robot $j$ that has a possible binding assignment containing at least one unique binding, i.e. robot $j$ contains at least one binding that can only be assigned to it. If multiple robots qualify, one is selected at random, and its unique bindings are stored in $r_j^*$ (line \ref{alg:alloc:line:unique}). During the original allocation process, the final assignment for robot $j$, $r^{new}_j$, is the largest set of bindings it can do that contains the bindings in $r_j^*$; for reallocation, preference is given to the original robot assignment $r_j$.
    \item If none of the unassigned robots satisfy the previous criteria, then we find the set $R'$, where robot $j$'s possible binding assignment set $R_j$ is in $R'$ if and only if robot $j$ can be assigned to at least one binding that is currently unassigned (line \ref{alg:alloc:line:unassigned}). If multiple robots qualify, the robot with the least number of elements in its possible binding assignment set $R_j$ (i.e. the robot with the least flexibility in its assignment) is chosen (line \ref{alg:alloc:line:choosemin}). We do this to ensure that the robot with the most flexibility in its assignment (i.e. has the most binding assignment options) will not be chosen first. Similar to before, during reallocation, preference is given to the original robot assignment $r_j$; otherwise, the final assignment for robot $j$, $r^{new}_j$, is the largest set of bindings it can do that contains a binding in the set of unassigned bindings (line \ref{alg:alloc:line:rnew2}).
    \item If all bindings have been assigned, robot $j$ is chosen at random. During the original allocation process, it is assigned the maximally-sized set of bindings (line \ref{alg:alloc:line:rnew3}); during reallocation, preference is given to the original robot assignment $r_j$.
\end{enumerate}


For the constraints $c_{min}$, after a robot is assigned a set of bindings $r_j^{new}$, we update $c_{min}$ to be the set $\{ (\rho, k-1) \ | \ (\rho, k) \in c_{min}, \rho \in r_j^{new}, k-1 > 0\}$ 
(line \ref{alg:alloc:line:updateC}). Intuitively, for each binding $\rho \in r_j^{new}$, we decrement the corresponding value of $k$, which represents the minimum number of robots that are still required to be assigned $\rho$. If $k-1 \leq 0$ (i.e. at least $k$ number of robots have now already been assigned to $\rho$), we remove $(\rho, k)$ from $c_{min}$ and remove $\rho$ from the set of bindings that have not been assigned yet (line \ref{alg:alloc:line:update_unassigned}).

Because the robot is assigned the largest set of bindings, the team may have overlapping assignments, i.e. robots can be removed while still ensuring the overall task will be completed, which is beneficial for robustness. However, the teaming assignment may vary depending on the ordering in which the robots are assigned. Thus, the reassignment may not be the globally optimal solution. 

\begin{algorithm}
    \SetKwInOut{Input}{Input}
    \SetKwInOut{Output}{Output}
    \SetKwProg{Initialization}{Initialization}{}{}
    \Input{$\mathbf{b}$, $\!R = \!\{R_1, ..., R_n\}$, \new{$\!\mathcal{R}_{A} \!=\! \{r_1, ..., r_n\}$}\!\!, \newc{$c_{min}$}\!\!\!} 
    \Output{$\mathcal{R}^{new}_{A}$}
    $unassigned = \mathbf{b}$ \\
    \While{$R \neq \emptyset$}{
    $r_j^{*} = \textsc{get\_unique\_r}(R)$ 
    \label{alg:alloc:line:unique}\\
    \If{$r_j^{*} \neq \emptyset$}{
        $R_j^{*} = \{r \in R_j \ |\ r_j^{*} \subseteq r\}$ \\
        $r^{new}_j = \begin{cases}
              \new{r_j} & \new{\text{if } r_j \neq \emptyset, r_j \in R_j^*} \\
              \textsc{randOf}(\text{argmax}_{r \in R_j^{*}} |r|) & \text{otherwise}
            \end{cases}   $
            
        }
    \Else{
    
    $R' = \{R_j \in R \ | \ \exists r \in R_j \text{ s.t. } r \cap unassigned \neq \emptyset \}$ \label{alg:alloc:line:unassigned} \\

    \If{$R' \neq \emptyset$}{
    $R_j = \text{argmin}_{\hat{R} \in R'} |\hat{R}|$ \label{alg:alloc:line:choosemin}\\
    $R'_j = \{r \in R_j \ | \ r \cap unassigned \neq \emptyset \}$\\
    $r^{new}_j = \begin{cases}
          \new{r_j} & \!\!\!\!\!\!\!\!\new{\text{if } r_j \neq \emptyset, r_j \in R'_j}\\
          \textsc{randOf}(\text{argmax}_{r \in R'_j} |r|) & \text{otherwise}
        \end{cases}   $ \label{alg:alloc:line:rnew2}
    
        }
    \Else{ \label{alg:alloc:line:random} 
         $r^{new}_j = \begin{cases}
              \new{r_j} & \!\!\!\!\!\!\!\!\new{\text{if } r_j \neq \emptyset, r_j \in R_j} \\
              \textsc{randOf}(\text{argmax}_{r \in R_j} |r|) & \text{otherwise}
            \end{cases}   $  \label{alg:alloc:line:rnew3}  
    }
    }

    $R \setminus R_j$,
    $\mathcal{R}^{new}_{A} \cup \{r^{new}_j\}$ \\
    \newc{$c_{min} = \textsc{update\_c}(c_{min}, r^{new}_j)$} \label{alg:alloc:line:updateC}\\
    \newc{$unassigned = \textsc{update\_unassigned}(unassigned, c_{min}, r^{new}_j)$\!\!\!} \label{alg:alloc:line:update_unassigned}
    }
    \If{$unassigned = \emptyset$}{
    \Return $\mathcal{R}^{new}_{A}$}
    \Else{
    \Return $\emptyset$ 
    }

\caption{Binding (Re)allocation}
\label{algo:allocation}
\end{algorithm}

\section{Demonstration and Evaluation}

We illustrate the modification resynthesis framework using the example in Sec. \ref{sec:ex}. The behavior of the robots as modifications occur in simulation is shown in the accompanying video. 

\subsection{Mod 1: Adding Transitions}

During execution, the blue robot gains the ability to move between rooms B and G (Figure~\ref{fig:cap_mod}(b)), $\Delta^{add}_{blue} = \{(\{roomG_c\}, \{roomG_c, roomB\}),$ $(\{roomG_c, roomB\}, \{roomB_c\})$, $(\{roomB_c\}, \{roomB_c, roomG\})$, $(\{roomB_c, roomG\}, \{roomG_c\})\}$. On a physical system, this could represent a door opening or a ramp being introduced between the two rooms. Because adding transitions does not violate the current behavior of the robot, the blue robot stores this modification and continues executing its original behavior. The overall time for this modification was 0.00715 ms.

\subsection{Mod 2: Removing Transitions without Reallocation}

The orange robot can no longer move between room D and the hall, $\Delta^{rem}_{orange} = \{(\{roomD_c\}, \{roomD_c, hall\})$, $(\{roomD_c, hall\}, \{hall_c\})$, $ 
(\{hall_c\}, \{hall_c, roomD\})$, $(\{hall_c, roomD\}, \{roomD_c\})\}$. This could represent a door closing, or the size of the entrance changing such that the robot is no longer able to move through it.

The orange robot's original behavior included transitioning from room D to the hall in order to get to room B. Thus, it updates its model and product automaton and checks if it can still satisfy its original binding assignment $r_{orange} = \{\RN{1}\}$. Since it can still reach room B by going through rooms D and C, the robot can still satisfy binding \RN{1}. Thus, the orange robot resynthesizes its behavior, and no other robots are affected. The overall time for this modification was 18.79 ms.

\subsection{Mod 3: Removing Transitions with Reallocation}

During execution, the pink robot's camera fails, $\Delta^{rem}_{pink} =$ $ \{(\emptyset, \{camera\}), $ $(\{camera\}, \{camera\})\}$. In this scenario, the robot is unable to perform its original binding assignment $r_{pink} = \{\RN{2},\RN{3}\}$; since it no longer has a camera, it cannot satisfy binding \RN{2}, and therefore $r_{fail} = \{\RN{2}\}$. If any other robots were already assigned binding \RN{2}, then reallocation is not required. However, this is not the case in the original assignment. Thus, the robots go through the reallocation process. After each robot updates their individual models and product automata, their possible binding assignments are $R_{green} = \{(\RN{1}), (\RN{2})\}$, $R_{blue} = \{(\RN{1})\}$, $R_{orange} = \{(\RN{1})\}$, $R_{pink} = \{(\RN{1}), (\RN{3}), (\RN{1}, \RN{3})\}$.

During reallocation, the green robot is reassigned from binding \RN{1} to binding \RN{2} and the pink robot is reassigned to bindings \RN{1} and \RN{3}; all other robots maintain their original assignment and therefore do not resynthesize their behavior. The overall time for this modification was 196.03 ms. The time for the pink robot to update its model and product automata and find $r_{fail}$ was 109.0 ms; subsequently, the time for the remaining robots to update their models and product automata was 21.35 ms; the time for the task reallocation was 0.0699 ms; the time for the robots to resynthesize their behavior was 37.39 ms.

\section{Conclusion}

We introduced a hierarchical method for a team of heterogeneous robots to react to modifications in their capabilities during execution of a \ltlpsi \ specification. 
We also increase the expressivity of the \ltlpsi \ grammar by allowing the user to require a minimum number of robots for a binding, as well as constrain which  bindings cannot be assigned to the same robot. We implemented our approach in simulation in a warehouse scenario.

In the future, we plan to extend the resynthesis framework to other aspects of reactivity, such as reacting to external events. We also plan to explore ways to incorporate optimality {(e.g. minimizing cost)}
when finding a teaming assignment, as well as relaxing all-to-all communication constraints as they synchronize their behavior. 

\bibliographystyle{ieeetr}
\bibliography{references}
\end{document}